\documentclass[conference]{IEEEtran}
\IEEEoverridecommandlockouts
\usepackage{cite}
\usepackage{amsmath,amssymb,amsfonts}
\usepackage{graphicx}
\usepackage{textcomp}
\usepackage{xcolor}
\usepackage{amsmath}
\usepackage{empheq}
\usepackage{amsfonts}

\usepackage{algorithm}
\usepackage{algpseudocode}
\usepackage{algpseudocode}

\def\BibTeX{{\rm B\kern-.05em{\sc i\kern-.025em b}\kern-.08em
    T\kern-.1667em\lower.7ex\hbox{E}\kern-.125emX}}
\begin{document}

\title{Genetically Modified Wolf Optimization with
Stochastic Gradient Descent for Optimising Deep
Neural Networks}
\thispagestyle{plain}
\pagestyle{plain}

\author{\IEEEauthorblockN{1\textsuperscript{st} Manuel Bradicic}
\IEEEauthorblockA{\textit{Department of Computer Science} \\
\textit{University of Surrey}\\
Guildford, United Kingdom \\
mb01761@surrey.ac.uk}
\and
\IEEEauthorblockN{2\textsuperscript{nd} Michal Sitarz}
\IEEEauthorblockA{\textit{Department of Computer Science} \\
\textit{University of Surrey}\\
Guildford, United Kingdom \\
ms02264@surrey.ac.uk}
\and
\IEEEauthorblockN{3\textsuperscript{rd} Felix Sylvest Olesen}
\IEEEauthorblockA{\textit{Department of Computer Science} \\
\textit{University of Surrey}\\
Guildford, United Kingdom \\
fo00150@surrey.ac.uk}
}
\maketitle

\begin{abstract}
When training Convolutional Neural Networks (CNNs) there is a large emphasis on creating efficient optimization algorithms and highly accurate networks. The state-of-the-art method of optimizing the networks is done by using gradient descent algorithms, such as Stochastic Gradient Descent (SGD). However, there are some limitations presented when using gradient descent methods. The major drawback is the lack of exploration, and over-reliance on exploitation. Hence, this research aims to analyze an alternative approach to optimizing neural network (NN) weights, with the use of population-based metaheuristic algorithms. A hybrid between Grey Wolf Optimizer (GWO) and Genetic Algorithms (GA) is explored, in conjunction with SGD; producing a Genetically Modified Wolf optimization algorithm boosted with SGD (GMW-SGD). This algorithm allows for a combination between exploitation and exploration, whilst also tackling the issue of high-dimensionality, affecting the performance of standard metaheuristic algorithms. The proposed algorithm was trained and tested on CIFAR-10 where it performs comparably to the SGD algorithm, reaching high test accuracy, and significantly outperforms standard metaheuristic algorithms.
\end{abstract}


\section{Introduction}
The purpose of this paper is to research and discuss the use of evolutionary algorithms for finding the optimal weights of NNs to perform image classification on the CIFAR-10 dataset,  proposed by Krizhevsky et al.~\cite{krizhevsky2009learning}. Section 2 begins with a review of the work relating to CNNs and their emergence, an evolutionary algorithm Social Learning Particle Swarm Optimisation (SL-PSO), and population-based meta-heuristic algorithms for NN weight optimization in general. The detailed structure of our NN and the justification of the chosen architecture are presented in Section 3.
Section 4 explores a detailed structure of the proposed hybrid GWO-SGD~\cite{MIRJALILI201446}. In Section 4 we explore the collected results. Section 5 reports the training of the proposed algorithm, as a bi-objective problem, and discusses its overall performance. Finally, section 6 gives the conclusion of this work.

\section{Related Work} \label{label:background}
\subsection{Convolutional Neural Networks}
A typical Feed-forward NN (FFNN), also referred to as a multilayer perceptron (MLP), is composed of an input layer, hidden layers and an output layer. Each individual layer of neurons is interconnected with its neighboring layers through a set of real-valued weights. Furthermore, each layer of neurons, apart from the input layer, is assigned an additional activation threshold, called a bias. In this work, CNNs are used. The main property of CNNs that make them more suitable than FFNNs for this problem is that the inputs are images, on which they can perform convolutions. The reason they are better equipped is due to the fact that the convolutional layers that a CNN has, can successfully capture the spatial and temporal dependencies of an image. Convolution is a linear operation that takes two functions $f$ and $h$ and produces another function $g$. In the case of CNNs, $f$ is a multi-dimensional array (image pixels), $h$ is a pre-defined matrix,  array (also called a kernel or filter) and finally, $g$ is the result of the convolution, also called a feature map.
FFNNs and CNNs have two main phases: feed-forward and back propagation, through which the continuous optimization of weights and biases occurs. During the training process it is crucial to choose a suitable optimizer. Generally, image classification is attempted with gradient descent methods (GDs) \cite{Huang_2017_CVPR, tan2021efficientnetv2}. Based on how well the network performed on the input data, the GD updates the parameters of the network with a cost function. The aim is to reach the minimum of that cost function by taking small steps in the direction of the negative gradient.  One drawback is their tendency to converge easily towards local optima \cite{XUE2022453}. 
SGD \cite{robbins1951stochastic} as one of the most favored gradient-based algorithms for training NNs, also suffers from early convergence. To avert premature convergence, a wide range of adaptive gradient algorithms have been developed that adjust the learning rate in efficient ways, one worth mentioning is Adam.
The issue of premature convergence has been tackled in past literature through trying to improve the global search of gradient descent methods \cite{NEURIPS2018_62da8c91, XUE2022453}. All of the proposed solutions in these respective papers focus on hybridizing the very effective convergence speed of gradient descent with the gradient-less global search of meta-heuristic optimization algorithms. This paper aims to expand on this method of training neural networks, by hybridizing GWO \cite{MIRJALILI201446, Amirsadri2018-yo} and SGD together to minimize loss. Some elements from genetic algorithms (GA)\cite{MCCALL2005205} to diversify GWO have been used as well.

\subsection{Training Algorithms}
Metaheuristic algorithms are a rapidly expanding field with many variants to fit different problems. The No Free Lunch theorem \cite{wolpert1997no} has logically proven that there is no metaheuristic algorithm that can solve all optimization problems. Therefore, the exploration of algorithm variants and hybridizations may prove fruitful in the problem of finding the optimal set of weights for a CNN working on CIFAR-10. As a branch of metaheuristic algorithms, swarm intelligence algorithms have proven quite successful in such a task \cite{albeahdili2015hybrid}. Therefore, the decision was made to apply the SL-PSO and GWO algorithms to this problem, to explore their efficacy in finding an optimal solution.
Being a variant of the original Particle Swarm Optimization algorithm (PSO) \cite{kennedy1995particle}, Social Learning Particle Swarm Optimization (SL-PSO) was a method proposed by Cheng and Jin \cite{CHENG201543} to incorporate the social learning aspects of animals into ordinary PSO. This was done with the purpose of changing the trial and error process of asocial learning to a more social process where individuals learn from all higher performing individuals (demonstrators) in their respective swarm. SL-PSO implements swarm sorting and behavior learning \cite{CHENG201543} where the swarm is sorted into fitness values and every particle but the best one will learn from their respective demonstrators. Each particle apart from the best is considered an imitator. Furthermore, each particle also gets updated by a random inertia component, which applies diversity to the swarm and improves on the global search capabilities of the standard PSO algorithm~\cite{ZHANG2019109}. SL-PSO was also designed to be able to efficiently optimize problems of a higher dimensionality~\cite{CHENG201543}, which made it a good candidate to be used as a baseline for a network with a large amount of parameters.

The GWO algorithm~\cite{MIRJALILI201446} was proposed by Mirjalili and took inspiration from the hierarchical nature through which wolves interact and hunt in the wild. This hierarchy consists of alpha, beta, delta (dominant) and omega wolves(non-dominant) classified from highest fitness to lowest respectively. This is mathematically represented in the algorithm through a method similar to the social aspect of SL-PSO where the non-dominant omega wolves learn from the mean value of the dominant wolves. Further explanation of the GWO implementation can be seen in section 4. Due to the similarities between SL-PSO and GWO in their social learning, a comparison of the two would provide useful insight into their relative impacts on the optimization problem. GWO is described as having a highly efficient global search and good ability to converge~\cite{MIRJALILI201446} whereas SL-PSO was lacking in terms of its exploitation ability~\cite{ZHANG2019109}. Therefore, GWO is a logical candidate to use as an implementation in this research.

\subsection{Population Meta-heuristics for Neural Networks}
As previously mentioned, there are a variety of metaheuristic algorithms to choose from. A study found that population-based metaheuristic algorithms can be more efficient than the exact optimization algorithms, as the latter struggles to solve problems in a high-dimensional space \cite{beheshti2013review}. Furthermore, the authors argue that those algorithms have to make different assumptions in order to perform well in each problem. Population-based algorithms step away from this issue as they usually make far less assumptions or no assumptions at all about the problem. As mentioned in the previous section, they can search large decision spaces and are not prone to getting stuck in local minima, unlike the gradient descent methods. A survey was done on evaluating the performance of training feedforward neural networks with the metaheuristic algorithms \cite{Fong16}. They analyzed various algorithms, and in some cases, they matched or even outperformed the gradient-based methods on low dimensional neural networks \cite{ mirjalili2015effective, mosavi2016classification}. Many population-based metaheuristics have a strong exploration and exploitation, whereas gradient-based methods only focus on exploitation. However, the problem comes when the metaheuristic algorithms are to be used for training deep neural networks which increase the number of decision variables from dozens up to even millions when considering modern architectures. 
A large part of the literature regarding the optimization of deep neural networks, and specifically CNNs, focus on optimizing the neural network architecture \cite{de2021gacnn, 9185541}. This approach generates the optimal networks for the given problem, however, this research aims to design an algorithm which trains the weights of the network. There were multiple instances in the literature~\cite{XUE2022453, albeahdili2015hybrid, NEURIPS2018_62da8c91} of different researchers combining gradient-based methods with metaheuristic algorithms. This was done to train neural networks by exchanging information between the two types. For instance, Albeahdili et al. \cite{albeahdili2015hybrid} created a hybrid of PSO and GA, and combined it with SGD, which demonstrated highly competitive results for CIFAR-10. The approach focuses on training the metaheuristic algorithm and then using SGD on those individuals. An issue with running multiple SGD algorithms is that on more complex networks it takes a very long time to run in parallel, and is computationally expensive. However, this approach has a strong benefit where the particles explore the space and then gradient descent is applied on them to exploit the space with various parameters. Furthermore, there is a huge advantage of using GAs to share information between different individuals, as it helps explore the search space more and helps guide the particles away from local minima \cite{10.2307/24939139}. The literature provided strong evidence that combining some population-based algorithms with genetic modifications yields better performance \cite{8329968, albeahdili2015hybrid, chen2009hybrid}.
Returning to \cite{albeahdili2015hybrid}, one drawback, as identified in the meta-analysis by \cite{Fong16}, is that even though PSO has a fast convergence rate, it has a low ability to find global minima. On the contrary, as mentioned in the previous section, GWO has a very high convergence rate as well as the ability to find global optima in high dimensions. Furthermore, as \cite{8329968} suggests, combining GWO with GA can improve on some of the shortcomings of the algorithm.

\subsection{Summary}
To conclude, this research proposes an algorithm which takes the advantage of combining a metaheuristic algorithm with a gradient descent method. This will allow it to explore the space, as well as exploit it better. Due to high dimensions presented by deep neural network models the exploitation part will be increasingly more difficult for metaheuristic algorithms as the model complexity increases. This algorithm will exchange information between the two training types in the form of individuals. To differ from the literature, this research proposes a hybrid between GWO and GA, and combines it with a gradient descent method, such as SGD. The choice to use SGD as opposed to Adam, is because converges faster and the adaptive feature of it won’t work very well when the network weights are constantly being switched \cite{kingma2014adam}. This proposed method is novel because there were no attempts to optimise deep neural network weights with this hybrid algorithm. To avoid premature convergence, GA modifications were proposed \cite{8329968,albeahdili2015hybrid}, by using crossover and mutation genetic operators only when the fitness stagnates.

\section{Selected Architecture and Justification}

\begin{figure*}
  \includegraphics[width=15cm, height=5cm]{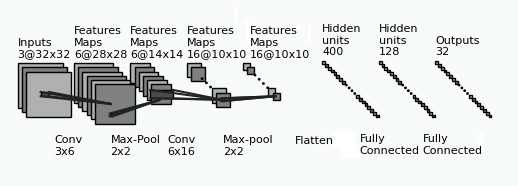}
  \caption{Selected Neural Network Architecture}
  \label{fig:architecture}
\end{figure*}

CNN architecture is another popular topic that has been in the limelight when it comes to optimization as it can be a long and difficult process to find the correct architecture that suits a specific problem \cite{khan2020survey}. After having reviewed a series of popular architectures \cite{karim_2022, khan2020survey}, it was found that even among the smallest ones, the number of features still ranged in the hundreds of thousands. This network size is very costly for the optimization problem of finding an optimal set of weights, as each feature is another decision variable that would need to be trained. To combat this computational difficulty, a smaller, custom network architecture was used for training against the CIFAR-10 dataset instead. This network has a total of 58,685 features. An illustration of the architecture can be seen in Fig. \ref{fig:architecture}. It is also worth mentioning that the option of transfer learning and block training was considered, through which a pre-trained model would be used to bolster the accuracy of the model. A few unfrozen layers would be appended to the frozen model for the training algorithm to optimize. These solutions would remove the issue of high-dimensionality, however, it would go against the purpose of this research by not using the global search capabilities of GWO in a meaningful way.

\section{Training Algorithm}

The implemented algorithm, GMW-SGD, is displayed in the pseudocode in Algorithm \ref{alg:pseudocode}. The proposed algorithm is a hybrid between GWO \cite{MIRJALILI201446} and SGD, with occasional mutation or crossover.

Initially, the population of individuals is initialized. As defined in \cite{MIRJALILI201446}, the individuals depict wolves in a pack. Each one of them represents the parameters from the CNN; either a weight or a bias \cite{ZHANG20071026}, and hence, the individual will be a flattened representation of the network. The encoding strategy becomes difficult for large CNNs with thousands of parameters and many layers, so the simplified encoding of an individual would be represented as follows:

\begin{equation} \label{eq:ind}
    ind(i) = [w_0, ... , w_{Nw},b_0, ... , b_{Nb}],
\end{equation}

\begin{equation} \label{eq:population}
    population = [ind(1);...; ind(Np)],
\end{equation}

where $Np$ is the size of population, $i$ = 1, …, $Np$, $Nw$ is the number of weights and $Nb$ is the number of biases in the network. The fitness of the individuals is calculated with Categorical Cross Entropy loss function shown in Eq. (\ref{eq:CE}) and the objective is to minimize it. As per original GWO algorithm, there is a social hierarchy of grey wolves represented mathematically that the algorithm follows. The three best individuals are saved in the descending fitness into Alpha ($\alpha$), Beta ($\beta$) and Delta ($\delta$) wolves, respectively, and they represent the $dominating$ wolves. Rest of the individuals are saved as Omega wolves ($\omega$).

\subsection{Grey Wolf Optimization Algorithm}

The basis of the GWO algorithm mimics the stages of grey wolf hunting in the wild. The algorithm is split into different stages: encircling the prey, hunting and attacking the prey.

During each iteration of the algorithm, all of the $omega$ wolves change their positions according to the positions of $dominant$ wolves. The assumption is that those three will have better knowledge about the potential location of prey, i.e. optimal solution. Therefore, the rest of the pack diverges from each other to search for the prey, with the use of $\vec{A}$ and $\vec{C}$ as shown in Eqs. (\ref{eq:A_const}) and (\ref{eq:C_const}), and then they converge to attack the prey when it is found. The locations of the $omega$ individuals are updated by the following formulas:

\begin{subequations}
  \begin{empheq}{align} \label{eq:D_dominant}
     \vec{D}_\alpha = |\vec{C_1} \cdot \vec{X}_\alpha(t) - \vec{X}(t)|, \\ 
     \vec{D}_\beta = |\vec{C_2} \cdot \vec{X}_\beta (t) - \vec{X}(t)|, \\
     \vec{D}_\delta= |\vec{C_3} \cdot \vec{X}_\delta(t) - \vec{X}(t)|,
  \end{empheq}
\end{subequations}

\begin{equation} \label{eq:X_dominant}
    \vec{X}_m(t+1) = \vec{X_l}(t) - A_m \cdot D_l,
\end{equation}

\begin{equation} \label{eq:X_t1}
\vec{X}(t+1)=\frac{(\vec{X}_1+\vec{X}_2+\vec{X}_3)}{3},
\end{equation}

where $t$ signifies the current generation, $\vec{X_l}$ is the position of the prey indicated by $l$ = \{$\alpha$, $\beta$, $\delta$\}. ($\vec{X}$) is the new location of the current wolf and $m$ = \{1,2,3\}. Finally, $\vec{A}$ and $\vec{C}$ are constant vectors calculated as follows:

\begin{equation} \label{eq:A_const}
\vec{A} = 2\vec{a}\vec{r_1} - \vec{a},
\end{equation}

\begin{equation} \label{eq:C_const}
\vec{C} = 2\vec{r_2},
\end{equation}

where $r_1 \text{,}r_2$ are randomly generated vectors in range [0,1] and the component $\vec{a}$ is linearly decreased over the generations from 2 to 0.

Despite the merits of GWO mentioned in \ref{label:background} Related Work, the algorithm is prone to converge prematurely and get stuck in local minima due to the high dimensionality. To tackle the issue, GA modifications will be applied.

\subsection{Genetic Modifications: Hybrid of GWO and GA}

By utilizing genetic crossover and mutation \cite{10.2307/24939139}, individuals can be modified when they stagnate. Therefore, the genetic algorithm is set to run only when $patience$ is reached, due to the fitness of the $omega$ wolves not improving. There is also a probability of whether the population will be mutated or crossed over with $p_{mut}$.

Firstly, to follow the algorithm, only the $omega$ wolves are mutated, as the $dominant$ wolves know where the prey is and they should not be altered. Each individual is mutated by the following formula:

\begin{subequations}
\begin{empheq}[left={p'}\empheqlbrace]{align}
   p + ((2u)^{\frac{1}{1+ \eta_m}}-1)(p-x_i^{(L)}), u \leq 0.5 \label{eq:mutation_1} \\ 
   p + (1 - (2(1-u))^{\frac{1}{1+ \eta_m}})(x_i^{(U)}-p), u > 0.5 \label{eq:mutation_2}
\end{empheq}
\end{subequations}
\vspace{0.1cm}

where $u$ is randomly generated in range from 0 to 1, $\eta$ is a constant in range [20,100], determining how similar to the parents the output will be, and the mutated individual ($p'$) is bounded by $x_i^{(L)}$ and $x_i^{(U)}$. The worse individuals are set to be mutated more, as they are performing worse.

Secondly, the crossover is carried between all the $omega$ wolves, where each is combined with a random $dominant$ wolf, as depicted in Figure 3. Similar to mutation, the rate of how many decision variables modified decreases with better individuals, so that the better individuals are not changed as much. The offspring take some parts from both of the parents, and in turn, $omega$ wolves will take some of the decision variables from the $dominant$ wolf which can help guide them better.

\begin{figure}[htp]
    \centering
    \includegraphics[width=6cm]{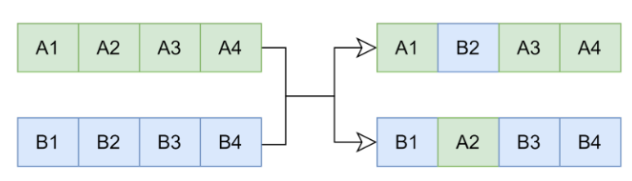}
    \caption{Crossover between $omega$ and $dominant$ wolf}
    \label{fig:crossover}
\end{figure}

To further tackle the issue of evolutionary algorithms struggling to exploit the high dimensional decision space, the above-mentioned metaheuristic algorithm was combined with a gradient descent method to help it exploit the space after exploration.

\subsection{Stochastic Gradient Descent}

Finally, this algorithm utilizes SGD, where it is run on the three best individuals ($\alpha,\beta,\delta$) after all the individuals are updated in the metaheuristic step. It fits naturally with the GWO algorithm as the best individuals are used to guide the rest of the pack. This simulates the $dominant$ wolves hunting the prey, as these ones are in the best locations out of the pack. The loss function used for the training was Categorical Cross Entropy ($CE$), as shown below:

\begin{equation} \label{eq:CE}
CE= -\log(\frac{e^{s_p}}{\Sigma^c_j{e^{s_j}}})
\end{equation}

where $C$ are the classes, $s$ is the score for a class and $p$ is the correct/positive class.

After the $dominant$ wolves are updated, the rest of the wolves will still explore the space, but over time will start moving towards the best ones. As they get closer they might find another prey (i.e. better minimum). Final optimization feature added to the algorithm was the slow decrease of the learning rate, implemented with the SGD to make the optimizer make smaller steps, to approach the minimum more.

The GMW-SGD combines the strong exploration introduced by GWO algorithm and is further enhanced by the genetic modifications to tackle some of the original GWO’s shortcomings. Finally, the information is exchanged between the evolutionary and the gradient descent part through the $dominant$ wolves, for which SGD is used to help with the exploitation of the best-found solutions.

\renewcommand{\algorithmicrequire}{\textbf{Input:}}
\renewcommand{\algorithmicensure}{\textbf{Output:}}

\begin{figure}[H]
    \includegraphics[scale=0.95]{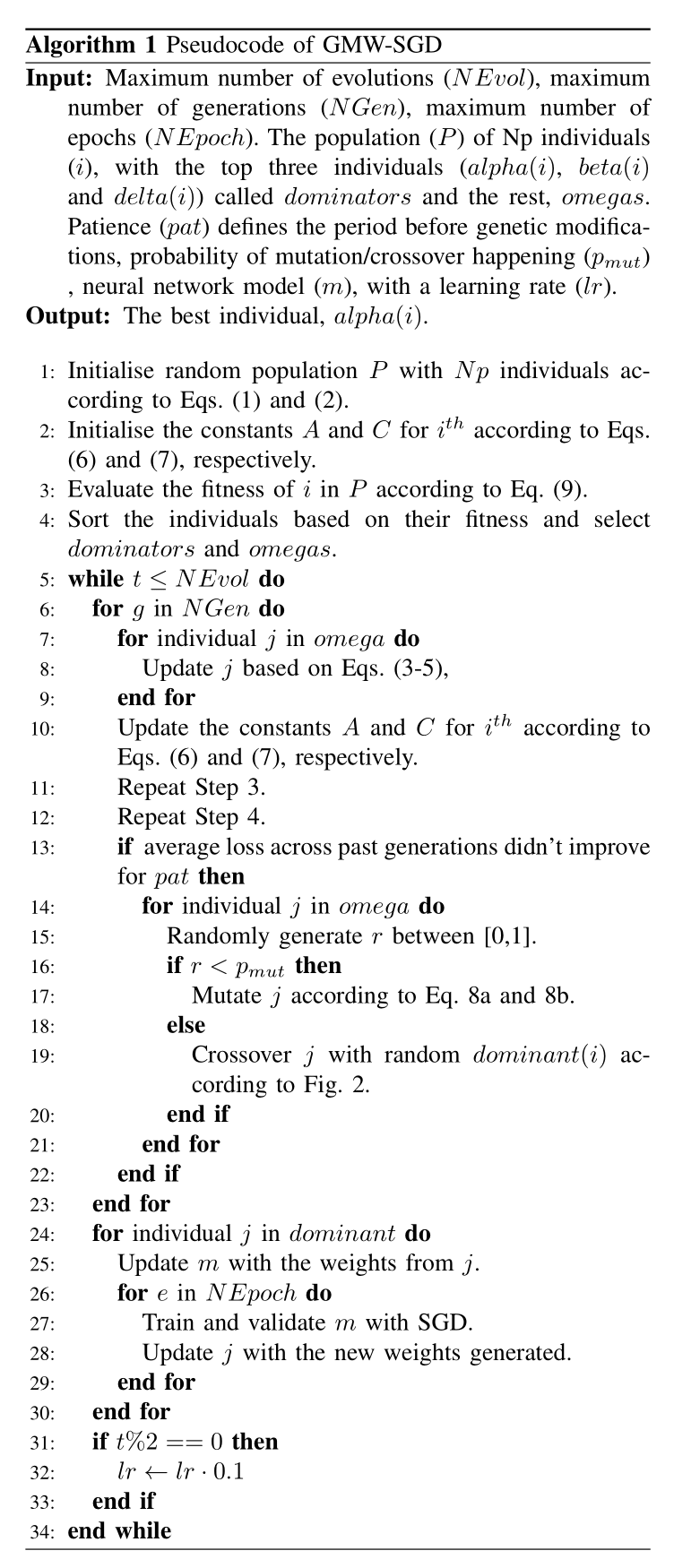}
    \label{alg:pseudocode}
\end{figure}

\section{Results}
The performance of the proposed algorithm was tested by benchmarking it with two other training algorithms. The selected algorithms are:
\begin{enumerate}
  \item SL-PSO - a population-based algorithm. It is closely related to the metaheuristic part of the algorithm.
  \item SGD - a gradient-based optimizer with a learning rate scheduler.
  
\end{enumerate}

The parameters that were used during the training process are shown in Table \ref{tab:parameters}.

\renewcommand\arraystretch{1.5}

\begin{table}[htp]
\centering
\caption{Parameters for the Algorithms}
\begin{tabular}[t]{lcc}
\hline
\textbf{Algorithm} & \textbf{Parameter} & \textbf{Value}\\
\hline
SGD & Learning rate ($\alpha$) & 0.01 \\
    & $\alpha$ Scheduler Factor & 0.1 \\
SL-PSO & Population size ($Np$) & 60 \\
       & Dimensions ($d$) & 58,685 \\
       & Number of Iterations ($NEvol$) & 36 \\
       & Position & [-0.1, 0.1] \\
       & Velocity & [-0.01, 0.01] \\
       & Constant $\alpha$ & 0.5 \\ 
       & Constant $\beta$ & 0.0001 \\
       & Constants $c1$ and $c2$ & 2 \\
GMW-SGD & $NEvol$ & 10 \\
        & $d$ & 58,685 \\
        & Epochs ($NEpoch$) & 2 \\
        & Generations ($NGen$) & 14 \\
        & $Np$ & 15 \\
        & Constant $a$ & [1,0] \\
        & Patience & 4 \\
        & Mutation/Crossover probability ($p_{mut}$) & 0.7 \\
        & Modification rate & [0.6, 0.1] \\
\hline
\end{tabular} \label{tab:parameters}
\end{table}

\begin{figure}[htp]
    \centering
    \includegraphics[width=9cm]{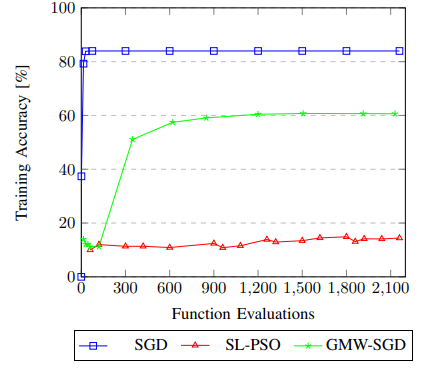}
    \caption{Convergence of best training accuracies on CIFAR-10}
    \label{fig:table_results}
\end{figure}

The algorithms were run once, independently, against CIFAR-10 and were added to Table \ref{tab:results_1} and Fig. \ref{fig:table_results}. Both of the metaheuristic algorithms were set to run for 2160 function evaluations. However, SGD ran for less as it quickly converged and got stuck in a minimum. Hence, it was stopped early, once it has not improved for about 20 epochs and made the assumption that the model would not change over time (and if it did it would overfit on the training dataset).

The performance of the models was evaluated by testing the accuracy on both training and test sets from CIFAR-10. Furthermore, CE was used to evaluate the loss on the test dataset.

\begin{table}[htp]
\caption{Results comparison of Algorithms on CIFAR-10}
\begin{center}
\begin{tabular}{|c|c|c|c|}
\hline
\textbf{Evaluated}&\multicolumn{3}{|c|}{\textbf{Evaluation Results}} \\
\cline{2-4} 
\textbf{Algorithms} & \textbf{\textit{Train Accuracy}}& \textbf{\textit{Test Accuracy}}& \textbf{\textit{CE}} \\
\hline
SGD & 83.97\% & 62.99\% & 1.3921 \\
SL-PSO & 14.42\% & 14.42\% & 2.2795 \\
GMW-SGD & 60.65\% & 60.63\% & 1.1170 \\
\hline
\end{tabular}
\label{tab:results_1}
\end{center}
\end{table}

After analyzing the results the following conclusions can be drawn. The first major difference that can be noticed in Fig. \ref{fig:table_results} is the difference between SL-PSO and SGD. This outlines the major difference between the two types of algorithms. SGD instantly exploits the decision space it starts in and successfully updates the weights to quickly reach higher accuracies. On the other hand, SL-PSO steadily increases over time, but the improvement it makes are extremely small. 

With GMW-SGD, it can be seen that the algorithm inherits trends from both types of algorithms. Unlike pure gradient descent methods, it takes longer to converge, and the individuals explore the space throughout the training to find the best solutions to exploit. It can be seen that it started converging about half way through the training, and even then it still increases by very little. Furthermore, it shows to be much better suited for this type of training than a standard population-based metaheuristic algorithm. SL-PSO might never reach high accuracy due to converging early, and if it did it would take a very long time. 

Even though the training accuracy of GMW-SGD might not reach as high as the SGD's, Table \ref{tab:results_1} suggests that SGD overfits substantially on the training data. The differences in the classifications on the test dataset were comparable between SGD and GMW-SGD, with the former performing just slightly better. Furthermore, the loss of GMW-SGD on the test dataset was smaller than the one of the model trained with SGD. These results suggest that the proposed algorithm can perform almost as well, if not just as well as a standard gradient-based method. As Table \ref{tab:results_1} suggests, the trained model transfer extremely well to the test dataset, thus not over-fitting at all. 

The proposed algorithm takes the best of both exploitation and exploration, and for that reason the performance is very similar. Thus, it was expected to perform better than SGD, but it could only match the metrics. One possible explanation why neither of the algorithms can perform better is due to the architecture being used. As mentioned previously, a simple network was used in order to make the dimensions smaller for metaheuristic algorithms. Most modern solutions use deeper and more complex neural networks for tasks such as this and the proposed neural network does not have the capacity to perform as well. Hence, it may affect the performance of both SGD and GMW-SGD. Secondly, even though the dimensions of the network are relatively small for the deep learning standards (like EfficientNet or MobileNet with over a million parameters), they are extremely high for metaheuristic algorithms, as it extensively increases the number of decision variables being changed. For this reason, the algorithm might not be able to perform as well, and therefore, might predominantly rely on SGD. If the decision space was smaller, the algorithm could potentially benefit more from the exploration of other individuals.

\section{Bi-Objective Problem}
Training as a bi-objective problem by non-dominated Sorting has also been demonstrated in this paper. Kalyanmoy et al. in \cite{996017} suggest a Non-dominated Sorting Genetic Algorithm II (NSGA-II) which diminishes the difficulties of the regular NSGA, such as computational complexity and nonelitism approach. In the proposed approach, once the offspring population has been created, the algorithm sorts both parent and offspring populations into different non-dominated fronts. This strategy is used to determine which individuals are to be chosen for the next generation. This bi-objective training problem has considered two objectives: the maximisation of accuracy, and the minimization of the Gaussian Regularizer (the sum of the square of the weights). The parameters used in this experiment remained the same as in the previous section, see Table I.

The overall observation is that the GMW-SGD optimizer algorithm did not perform successfully when training as a bi-objective problem due to several reasons. The problem of image classification that is trying to be solved in this study, prioritises high accuracy over the total sum of the network (the sum of the weights). The non-dominated sorting algorithm, due to its nature, would abolish elitism, the individuals with high accuracy and higher Gaussian regularisation value, prioritising less relevant individuals. GWO is a social learning algorithm that relies on the 3 most successful individuals who are ‘leading’ the pack. In this case, those individuals were not the best representation of the most successful wolves and were often individuals with a low Gaussian Regulariser value converging in a local minimum. This type of selection prevented the individuals from learning from the better ones and prevented the algorithm from reaching a global minimum, but rather converged in one of the local minima. In essence, this optimization problem is not suited for bi-objective optimization and performs significantly better as a single optimization problem. Table III displays a set of individuals with the highest crowding distances, and they are the most representative Pareto-optimal front.

\begin{table}[htbp]
\caption{Selected Individuals from the Pareto Front in a Bi-Objective Optimization}
\begin{center}
\begin{tabular}{|c|c|c|}
\hline
\textbf{Pareto Front}&\multicolumn{2}{|c|}{\textbf{Evaluation Metrics}} \\
\cline{2-3} 
\textbf{Individuals} & \textbf{\textit{Accuracy}}& \textbf{\textit{Gaussian Regularizer}}\\
\hline
Individual 1& 10.69 & 7.4533 \\
Individual 2& 24.85 & 11.7551 \\
Individual 3& 23.49 & 11.3150 \\
Individual 4& 11.14 & 7.6071 \\
Individual 5& 16.28 & 9.8632 \\
Individual 6& 13.4 & 9.3054 \\
\hline
\end{tabular}
\label{tab2}
\end{center}
\end{table}

\section{Conclusion and Future work}
Even though there exists a number of classical meta-heuristic optimization techniques, it can be concluded that they do not converge as efficiently as gradient-based methods. However, with ample computational resources, there is potential for them to reach the same optima. In this work, a hybrid evolutionary algorithm, combined with SGD, was proposed, utilizing a novel implementation of mutation that leads to the heterogeneous behaviour of particles across space. GMW-SGD brings together both exploration and exploitation and was able to perform better than the baseline meta-heuristic algorithm, while also matching the accuracy of standard SGD. On top of this GMW-SGD provides the added benefits of an improved population-based global search which improves the avoidance of local minima.

Some future improvements to this research can be made by running each algorithm for a larger amount of function evaluations. As the meta-heuristic algorithms would be given more chances to demonstrate their global search capabilities once the gradient-descent method has converged into an optimum. 

Further randomness could be implemented in addition to GA mutation and crossover such that the steps taken by the omega wolves in the direction of the mean of the dominant wolves are more varied. This should allow for further exploration in a less systematic way compared to the standard implementation of GMW-SGD.

To tackle the issue of optimizing weights for deep neural networks using metaheuristic algorithms, the training process could be broken down into training blocks. That means that the training process would consist of training weights per layer, rather than the network as a whole. That property would allow GWO-SGD to train on deep neural networks while still maintaining low dimensionality.

\bibliographystyle{unsrt}
\bibliography{bibliography.bib}

\end{document}